\documentclass{article}

\usepackage[preprint]{neurips_2026}

\usepackage[utf8]{inputenc} 
\usepackage[T1]{fontenc}    
\usepackage{hyperref}       
\usepackage{url}            
\usepackage{booktabs}       
\usepackage{amsfonts}       
\usepackage{nicefrac}       
\usepackage{microtype}      
\usepackage[table]{xcolor}         
\usepackage{graphicx}
\usepackage{enumerate}
\usepackage{enumitem}
\usepackage{float}
\usepackage{xspace}
\usepackage{xcolor}
\usepackage{listings}
\usepackage{wrapfig}

\lstdefinestyle{promptstyle}{
  basicstyle=\ttfamily\scriptsize,
  columns=fullflexible,
  breaklines=true,
  breakatwhitespace=false,
  keepspaces=true,
  showstringspaces=false,
  numbers=left,
  numberstyle=\tiny\color{gray},
  stepnumber=1,
  numbersep=6pt,
  frame=single,
  framerule=0.25pt,
  rulecolor=\color{black!35},
  xleftmargin=1.0em,
  xrightmargin=0.5em,
  aboveskip=0.75em,
  belowskip=0.75em,
  tabsize=2,
  captionpos=b,
  literate=
    {→}{{$\to$}}1
    {←}{{$\leftarrow$}}1
    {≥}{{$\ge$}}1
    {≤}{{$\le$}}1
    {§}{{\S}}1
    {—}{{---}}1
    {–}{{--}}1
    {“}{{``}}1
    {”}{{''}}1
    {’}{{'}}1
    {×}{{$\times$}}1
}

\definecolor{darkblue}{rgb}{0, 0, 0.5}
\hypersetup{colorlinks=true, citecolor=darkblue, linkcolor=darkblue, urlcolor=darkblue}

\usepackage{tcolorbox}
\tcbuselibrary{skins}
\tcbset{
  aibox/.style={
    width=\linewidth,
    top=8pt,
    bottom=4pt,
    colback=blue!6!white,
    colframe=black,
    colbacktitle=black,
    enhanced,
    center,
    attach boxed title to top left={yshift=-0.1in,xshift=0.15in},
    boxed title style={boxrule=0pt,colframe=white},
  }
}
\newtcolorbox{AIbox}[2][]{aibox,title=#2,#1}

\tcbset{
  dcbox/.style={
    width=\linewidth,
    top=8pt,
    bottom=4pt,
    colback=red!6!white,
    colframe=black,
    colbacktitle=black,
    enhanced,
    center,
    attach boxed title to top left={yshift=-0.1in,xshift=0.15in},
    boxed title style={boxrule=0pt,colframe=white},
  }
}
\newtcolorbox{DCbox}[2][]{dcbox,title=#2,#1}

\usepackage{tabularx}
\tcbset{
  findingsbox/.style={
    width=\linewidth,
    top=6pt,
    bottom=5pt,
    left=7pt,
    right=7pt,
    colback=blue!4!white,
    colframe=black,
    colbacktitle=black,
    coltitle=white,
    fonttitle=\bfseries,
    enhanced,
    attach boxed title to top left={yshift=-0.1in,xshift=0.15in},
    boxed title style={boxrule=0pt,colframe=black},
  }
}
\newtcolorbox{FindingsBox}[2][]{findingsbox,title=#2,#1}

\newcommand{\findingtile}[2]{%
  \textbf{#1}\par\vspace{1pt}{\raggedright #2\par}%
}

\let\oldparagraph\paragraph
\renewcommand{\paragraph}[1]{\vspace{-.6em}
\oldparagraph{#1}
}

\title{Which Models Are Our Models Built On?\\Auditing Invisible Dependencies in Modern LLMs}

\newcommand{\huggingface}{\raisebox{-1.5pt}{\includegraphics[height=1.05em]{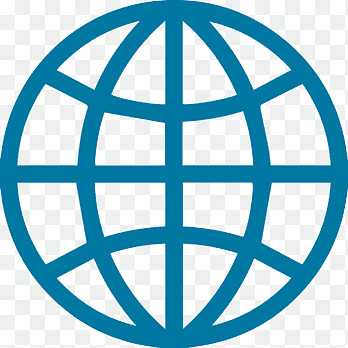}}\xspace}
\newcommand{\github}{\raisebox{-1.5pt}{\includegraphics[height=1.05em]{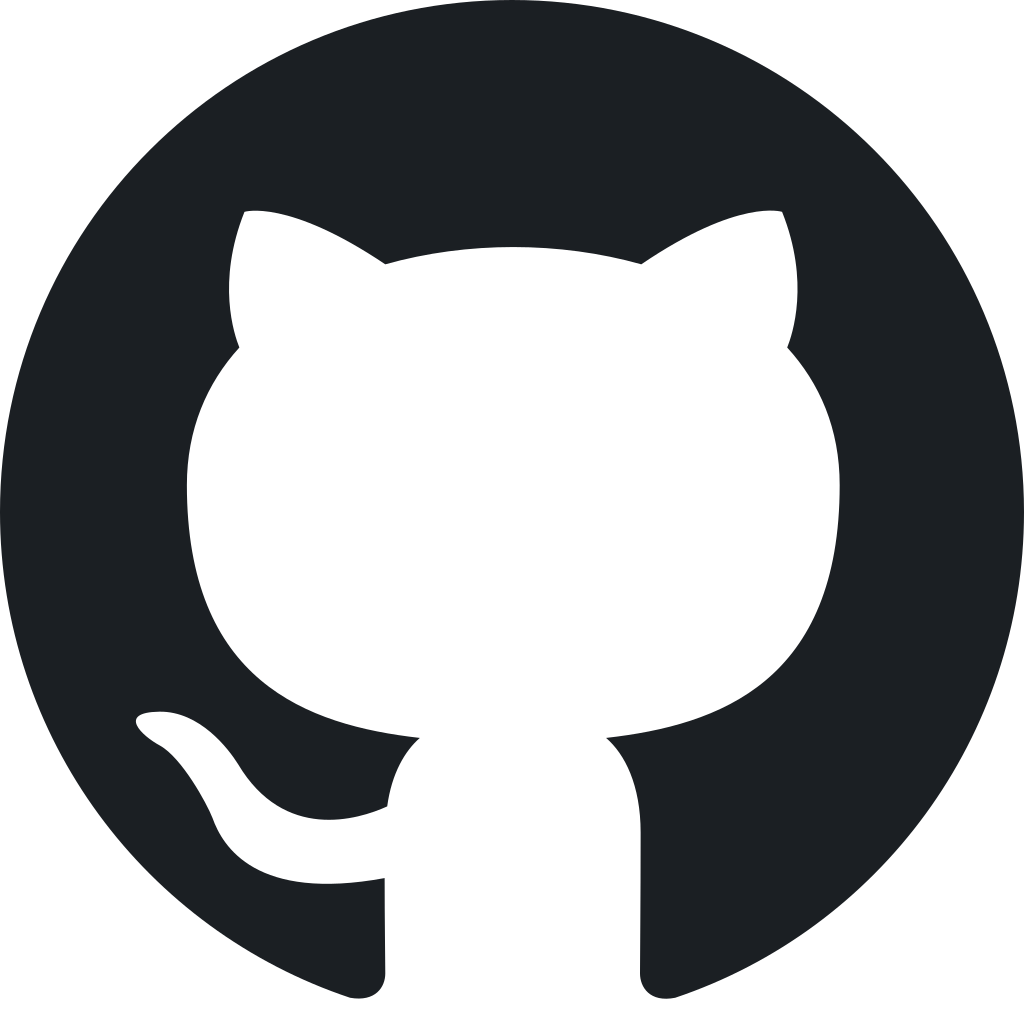}}\xspace}

\usepackage{pifont}
\newcommand{\system}{ModSleuth}

\newcommand{\myskip}[1]{}
\begin{document}

\makeatletter
\newcommand*\myfontsize{%
  \@setfontsize\myfontsize{9}{10}%
}
\makeatother

\author{%
\textbf{Sanjay Adhikesaven}\thanks{Equal contribution.}\hspace{0.4em}$^{1,2}$ \quad \textbf{Haoxiang Sun}$^{* 1}$ \quad \textbf{Sewon Min}$^{1, 2}$
\\[1ex]
$^1$University of California, Berkeley\quad$^2$Allen Institute for AI\quad
\vspace{.4em} \\
  \texttt{\{sanjay.adhikesaven, hxiang.sun, sewonm\}@berkeley.edu}\\
\vspace{-0.7em} \\
\github \textbf{Code}: \url{https://github.com/cal-data-audit/modsleuth}
\vspace{0.1em} \\
\huggingface \textbf{Demo}: \url{https://modsleuth.cal-data-audit.org}\\
}

\maketitle

\begin{abstract}
Modern LLM training pipelines increasingly rely on other models to generate data, filter corpora, judge outputs, and guide development decisions. These dependencies are \emph{recursive}: a model may depend on an upstream artifact whose own dependencies are documented only in separate releases and artifacts. As a result, the full dependency structure is fragmented across heterogeneous public artifacts, with complexity and recursive depth far outpacing humans' ability to trace. 

We introduce \textbf{\system{}}, an agentic system that recursively reconstructs LLM dependency graphs from public artifacts with source-grounded evidence. We find that the primary challenge is no longer information extraction, but defining what constitutes a dependency and reconciling artifact references across inconsistent documentation. We address these challenges through a formalization that distinguishes direct and indirect dependencies, represents heterogeneous pipeline roles through operation-centered relationships, and resolves artifact identities across names, versions, and repositories.

Applying \system{} to four public-artifact-rich LLM releases, we recover 1,060 source-verified dependencies and construct large-scale dependency graphs of modern LLM development. These graphs reveal multi-hop license obligations, train--evaluation coupling, discrepancies between released and training-time artifacts, and documentation inconsistencies that would otherwise be difficult to uncover. We release \system{} and the resulting dependency graphs to support transparent analysis of the increasingly complex ecosystems underlying modern LLMs.

\end{abstract}

\section{Introduction}

Modern large language models (LLMs) are increasingly shaped by other models in highly diverse ways---including data generation, rewriting, filtering, evaluation, preference learning, and other stages of development---rather than from raw human data alone
~\cite{wang-etal-2023-self-instruct, DBLP:conf/iclr/XuSZG0FTLJ24, DBLP:conf/icml/CuiY0YH0NXXL0024, DBLP:journals/corr/abs-2306-02707, DBLP:journals/corr/abs-2306-11644,liu-etal-2023-g,DBLP:conf/nips/ZhengC00WZL0LXZ23}.
As a result, LLM development has become deeply recursive: a model may depend on upstream artifacts whose own dependencies are documented only across technical reports, model cards, repositories, and datasets. These dependency chains are often fragmented and inconsistently documented, outpacing humans' ability to trace them manually---even for the original model creators themselves. 
For example, tracing dependencies of Olmo 3~\cite{DBLP:journals/corr/abs-2512-13961} requires first identifying upstream artifacts scattered across its technical report, model cards, and code repositories---including OCR systems, rewriting models, preference-learning pipelines and synthetic datasets---and then recursively repeating the same process for each upstream artifact.

This lack of transparent dependency structure has concrete consequences for \emph{responsible model and data use}. License restrictions may propagate silently through upstream synthetic datasets~\cite{longpre2024large, kim2025trustlicensesseedataset, jewitt2026permissivewashingopenaisupply}, data contamination can cascade through multi-hop paths that standard decontamination cannot trace~\cite{sainz-etal-2023-nlp, DBLP:journals/corr/abs-2311-04850}, and evaluations risk circularity when judge models share ancestry with the systems they evaluate~\cite{DBLP:conf/nips/PanicksseryBF24, DBLP:journals/corr/abs-2502-01534}.

We argue that this lack of transparency is \textbf{not incidental}, but \textbf{structural}: modern LLM development has evolved far faster than existing documentation and auditing efforts~\cite{DBLP:journals/tmlr/BommasaniKKLXML25, 10.5555/3716662.3716680}. 
Existing disclosure mechanisms (e.g., model cards, datasheets, and data cards~\cite{DBLP:conf/fat/MitchellWZBVHSR19, DBLP:journals/cacm/GebruMVVWDC21, DBLP:conf/fat/PushkarnaZK22}) provide useful schemas, but are often incomplete~\cite{DBLP:journals/corr/abs-2502-04484, liang2024systematic, DBLP:conf/iclr/YangL024} and fundamentally too flat to capture recursive, multi-stage dependencies.
Existing auditing approaches (e.g., ecosystem mapping~\cite{10.5555/3716662.3716680, DBLP:conf/cikm/RahmanGJ25, DBLP:journals/corr/abs-2502-04484, oderinwale2025anatomy, horwitz2025we}, ancestry inference from weights or behavioral signals~\cite{DBLP:conf/iclr/HorwitzSH25, wu2026llm, DBLP:conf/iclr/YaxOP25, 10.5555/3780338.3783543, kuditipudi2026blackbox, model_provenance_kit}, and dataset provenance tracing~\cite{li2026tracing}) similarly focus on narrow notions of lineage such as initialization or training data, and do not capture the diverse ways upstream models shape downstream LMs.

\begin{figure}
    \centering
    \includegraphics[width=1\linewidth]{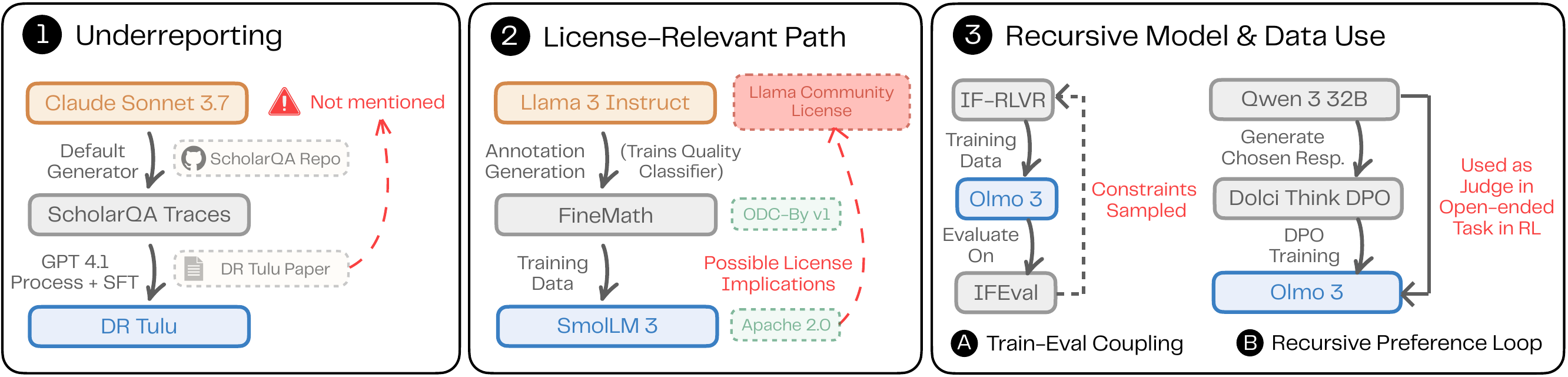}
    \caption{Dependencies \system{} surfaced: (1) DR Tulu's SFT traces to Claude Sonnet 3.7 via ScholarQA. (2) SmolLM3's FineMath traces back to a Llama-licensed artifact through a Llama-trained classifier. (3) Olmo 3 trains on IFEval-derived data while evaluating on it; Qwen3 32B serves as both DPO generator and RL judge.}
    \label{fig:placeholder}
\end{figure}

To address this gap, we introduce \textbf{\system}, an agentic system that recursively reconstructs LLM dependency graphs from public artifacts.
We find that, with recent advances in agentic capabilities (e.g., Claude Code \cite{claude_code_docs}), information extraction is no longer the primary challenge.
Instead, the key obstacles are semantic and representational: determining what constitutes a dependency, and resolving artifact references across inconsistent names, versions, model families, development stages, and repositories. Modern pipelines contain many ambiguous cases---including reward models, judge models, filtering classifiers, and model-generated datasets---whose influence ranges from directly affecting model weights to shaping development decisions without entering training.

We address these challenges through a formalization of recursive dependency tracing. Our framework distinguishes direct dependencies, which materially affect model weights, from indirect dependencies, which influence development without entering training. It further represents dependencies through operation-centered relationships (e.g., generation, filtering, rewriting, OCR, and evaluation) and introduces an identity lattice for reconciling references across heterogeneous sources.

The resulting graphs reveal an LLM ecosystem that is far more interconnected and compositional than previously recognized~\cite{DBLP:conf/cikm/RahmanGJ25, 10.5555/3716662.3716680, oderinwale2025anatomy}. Dependency chains extend up to eight hops and encompass a broad range of roles beyond model initialization~\cite{oderinwale2025anatomy} or dataset derivation and reuse~\cite{li2026tracing}, including synthetic data generation and curation~\cite{wang-etal-2023-self-instruct,DBLP:journals/corr/abs-2306-11644}, rewriting~\cite{DBLP:conf/iclr/XuSZG0FTLJ24}, data annotation~\cite{DBLP:conf/icml/CuiY0YH0NXXL0024}, OCR pipelines~\cite{DBLP:conf/iclr/BlecherCSS24}, and even the training data used for each of these auxiliary components. These graphs also surface several concerning practices
(Figure~\ref{fig:placeholder}): (1)~underreporting and inconsistencies across artifacts ({e.g.}, discrepancies between the paper and code), (2)~potential license and terms-of-use concerns arising from opaque reuse, and (3)~risks of contamination and evaluation bias introduced through recursive data generation and LLM-based evaluation.
In several cases, the recovered dependencies were unknown even to the original developers, suggesting that these issues stem not only from individual oversights but from structural limitations in current development and auditing practices.

Together, our work reveals a modern LLM ecosystem that is far more interconnected and recursively dependent than commonly understood, and provides a foundation for auditing, understanding, and governing the increasingly recursive ecosystems underlying modern AI systems.

\begin{AIbox}{Disclosure}
    \system\ identifies \textbf{\em declared} dependencies rather than true dependencies, meaning that the dependencies we detect likely represent only a lower bound on the actual dependencies. Importantly, comparisons across models, including intermediate models such as Qwen3, should be interpreted with caution, as a model may appear to have fewer or no dependencies simply due to limited public disclosure.
\end{AIbox}

\section{Background \& Related Work}

\oldparagraph{Background: Foundation Model Training.}
The past few years have seen the rapid consolidation of a foundation model ecosystem, with widely used artifacts spanning \emph{fully closed systems} (e.g., OpenAI~\cite{openai2026gpt55},
Gemini~\cite{googledeepmind2026gemini31pro},
Anthropic~\cite{anthropic2026opus47}), \emph{partially open models} accompanied by tech reports but limited transparency (e.g., Llama~\cite{DBLP:journals/corr/abs-2407-21783}, DeepSeek~\cite{DBLP:journals/corr/abs-2512-02556}, 
Qwen~\cite{DBLP:journals/corr/abs-2505-09388}),
and \emph{fully open-source efforts} (e.g., Olmo~\cite{DBLP:journals/corr/abs-2512-13961},
Nemotron~\cite{DBLP:journals/corr/abs-2512-20856},
Marin~\cite{hall2025marin}).

Historically, LLM training pipelines were relatively simple: large-scale pretraining on web corpora followed by post-training using curated human annotations. However, this paradigm has shifted toward increasingly complex, multi-stage pipelines in which models themselves play central roles. Upstream models are now routinely used to preprocess data (e.g., OCR~\cite{DBLP:conf/iclr/BlecherCSS24}), generate synthetic instructions or training data~\cite{wang-etal-2023-self-instruct, DBLP:journals/corr/abs-2306-11644}, produce answers and reasoning traces~\cite{DBLP:journals/corr/abs-2306-02707}, rewrite~\cite{DBLP:conf/iclr/XuSZG0FTLJ24} or filter data~\cite{DBLP:conf/nips/PenedoKALMRW024}, provide preference signals~\cite{DBLP:conf/icml/CuiY0YH0NXXL0024}, and serve as evaluators~\cite{liu-etal-2023-g, DBLP:conf/nips/ZhengC00WZL0LXZ23}.
This gives rise to a new class of \textbf{model-model dependencies}, where one model directly shapes another model artifact.
Such dependencies are highly diverse, complex, deeply recursive, and remain poorly visible---even when individually documented---because relevant information is scattered across sources and lacks a unified reporting scheme.
Despite their growing prevalence, their implications remain poorly understood: they introduce emerging risks of model collapse~\cite{DBLP:journals/nature/ShumailovSZPAG24}, bias reinforcement~\cite{DBLP:conf/nips/PanicksseryBF24, DBLP:journals/corr/abs-2502-01534}, unexpected behavioral transmission through training data~\cite{cloud2025subliminallearninglanguagemodels}, and contamination~\cite{DBLP:journals/corr/abs-2311-04850, sainz-etal-2023-nlp}, while also increasing the likelihood of unexamined license and terms-of-use implications~\cite{longpre2024large, kim2025trustlicensesseedataset, jewitt2026permissivewashingopenaisupply} and complicating attribution, provenance, and governance.

In this work, motivated by the emergence of deeply complex and recursive model-model dependencies, we formalize the task of recursive LLM dependency tracing and present \system{}, a system that extracts such dependencies from publicly available sources\footnote{
    In this work, we restrict our analysis to {\em reported} information from official sources, including technical reports, Hugging Face pages, and code releases, and therefore likely significantly underestimate the true extent of model-model dependencies, many of which are likely undocumented. Inferring such unreported dependencies is an important direction for future work.
} and makes them explicit for a given LLM. \system{} reveals deeply recursive structures that are otherwise difficult to find (e.g., hundreds of upstream artifacts for Olmo 3), and surfaces concerning implications, such as possible license and terms-of-use concerns and challenges for provenance (\S\ref{sec:findings}).

\paragraph{Related Work: Auditing ML Artifacts.}
A substantial body of work has sought to improve transparency and auditing of ML artifacts. Documentation frameworks such as model cards, datasheets, and data cards~\cite{DBLP:conf/fat/MitchellWZBVHSR19, DBLP:journals/cacm/GebruMVVWDC21, DBLP:conf/fat/PushkarnaZK22} promote structured reporting, but are often incomplete or inconsistently adopted~\cite{DBLP:journals/tmlr/BommasaniKKLXML25, DBLP:journals/corr/abs-2502-04484, liang2024systematic, DBLP:conf/iclr/YangL024}. Beyond self-disclosure, prior work has mapped ML ecosystem through metadata analysis and manual curation~\cite{10.5555/3716662.3716680, DBLP:conf/cikm/RahmanGJ25, DBLP:journals/corr/abs-2502-04484, oderinwale2025anatomy, horwitz2025we}, inferred model ancestry from weights or behavior~\cite{DBLP:conf/iclr/HorwitzSH25, wu2026llm, DBLP:conf/iclr/YaxOP25, 10.5555/3780338.3783543, kuditipudi2026blackbox, model_provenance_kit}, traced dataset provenance~\cite{li2026tracing}, and detected LLM-generated content in downstream artifacts~\cite{DBLP:conf/uss/00010S0025}.

These approaches largely focus on narrow notions of dependency such as weight initialization, fine-tuning, and dataset reuse. 
Many dependencies now central to LLM development---filtering, OCR preprocessing, rewriting, evaluation judging, and synthetic data generation---leave little or no trace in model parameters and are therefore difficult, if not impossible, to recover without explicit disclosure.
We instead reconstruct \emph{declared} dependencies from public artifacts, enabling recursive tracing of heterogeneous, multi-stage relationships that existing approaches largely miss. The resulting graphs should be viewed as an evidence-grounded lower bound on the true dependency structure, complementary to parameter-based inference methods.

\section{Design of \system{}}
\label{sec:method}

Given a target LLM release, our goal is to reconstruct an \emph{evidence-grounded dependency graph} over model and dataset artifacts using only public release evidence. Nodes correspond to model or dataset artifacts, and edges describe evidence-backed relationships through which an upstream artifact shapes a downstream artifact.

Surprisingly, we find that information extraction is no longer the primary bottleneck. Modern agentic systems such as Claude Code~\cite{claude_code_docs} can already navigate and synthesize complex technical documentation, and \system{} therefore uses Claude Code as its underlying extraction engine. The harder challenges are instead \textbf{semantic} and \textbf{representational}: defining what constitutes a dependency and resolving artifact identities across inconsistent names, versions, model families, development stages, and repositories. These challenges arise because modern LLM pipelines contain many ambiguous artifacts---including reward models, judge models, filtering classifiers, and model-generated datasets---whose influence ranges from directly affecting model weights to indirectly shaping development decisions. This section presents the key insights that address these challenges (\S\ref{subsec:task-def}), then describes the end-to-end \system{} pipeline (\S\ref{subsec:system-design}).

\subsection{Defining the Dependency-Tracing Task}
\label{subsec:task-def}

Modern LLM development relies on other models in increasingly diverse ways, making the notion of dependency fundamentally ambiguous. Upstream models may contribute through data generation, rewriting, filtering, evaluation, synthetic supervision, or more indirect forms of influence that resist clean attribution. Moreover, these operations are highly specialized, context-dependent, and rarely standardized, making fixed taxonomies insufficient.

\paragraph{What counts as a dependency.}
Some dependencies are straightforward: a model may be initialized from another model or trained on data generated by it.
However, modern LLM development has evolved to rely on upstream models in far more diverse ways.
For example, OCR systems and filtering models shape training data without generating it directly, and these components themselves depend on upstream models and datasets. At the same time, some artifacts influence development without entering training at all, such as evaluation models, ablation studies, or methodological recipes. The boundary becomes even less clear when influence is purely conceptual, such as related work citations.

To address this ambiguity, we distinguish between \textbf{direct} and \textbf{indirect} dependencies. A direct dependency is any upstream artifact that affects the target model's training or weights, including initialization models, synthetic-data generators, OCR systems, and filtering models. Direct dependencies are recursive: direct dependencies of direct dependencies are also considered direct dependencies, e.g., the data used to train an OCR or filtering model.

An indirect dependency includes an upstream artifact that does not directly enter training, but nevertheless substantially influences development decisions.
Examples include evaluation models or ablation variants that inform decisions, and methodological recipes explicitly adopted from prior work. In contrast, artifacts that neither affect training nor materially influence development---such as baseline comparisons, general related-work citations, or vague inspiration (e.g., ``following common practice'')---are excluded from the dependency graph.

\paragraph{How dependencies are represented.}
The roles played by upstream artifacts are too diverse to be captured by a fixed vocabulary of dependency types: many dependencies are highly specialized and nearly unique.
For example, the same model may be used to regenerate math problems, generate preference-learning completions, and evaluate outputs, each corresponding to a distinct relationship. Moreover, multiple upstream models often participate in the same pipeline stage with different responsibilities.

We therefore represent dependencies as \emph{operations}: structured groups of edges describing a single pipeline event.
Rather than solely relying on fixed dependency labels, each edge stores (1) a free-form natural-language description of how the dependency arises, (2) a coarse dependency type label used primarily for analysis, and (3) supporting source excerpts for verification.
When the same artifact participates in multiple stages, it appears in separate operations with distinct descriptions, preserving the full structure of its involvement.

\paragraph{Resolving artifact identity.} Finally, a common challenge is resolving artifact identity. Public sources often refer to the same model or dataset at different levels of specificity, and these references are frequently incomplete or ambiguous. A paper may mention a model family such as ``Olmo 3 32B'', while code points to a specific checkpoint or repository identifier; in some cases, even determining which concrete release is intended (e.g., \texttt{OLMoE-1B-7B-0924} vs.\ \texttt{OLMoE-1B-7B-0125}) requires substantial external context. Naively merging such references discards important uncertainty, while treating them as distinct artifacts fragments the dependency graph.

The problem is even more pronounced for datasets. Public artifacts often reference subsets, mixtures, derived variants, or internal names that do not map cleanly to canonical dataset identifiers. For example, establishing that \texttt{infiwebmath-3plus} is derived from \texttt{FineMath} requires tracing metadata across external repositories. As a result, identity resolution becomes a central challenge rather than a simple normalization step.

To represent this uncertainty, we organize artifact identity as an \emph{identity lattice}. Each artifact family contains a root node for underspecified references, intermediate nodes for partially resolved identities, and canonical leaves anchored to concrete URLs or release identifiers when available. Identity itself is represented as an open-vocabulary set of facets, such as \texttt{\{family:Olmo 3, size:32B, stage:Think\}}. This structure allows dependency claims to attach at the most specific level justified by the evidence, preserving uncertainty without forcing premature resolution.

\subsection{Full \system{} Design}
\label{subsec:system-design}

\begin{figure}
    \centering
    \includegraphics[width=1\linewidth]{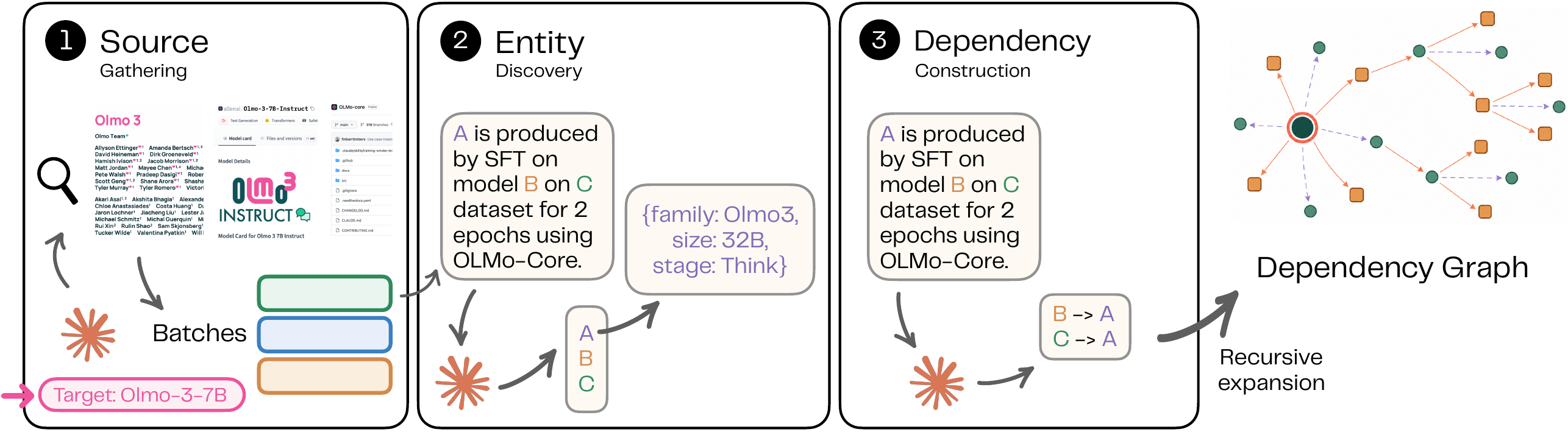}
    \caption{Overview of \system{}. \ding{202}~Public artifacts for the target release are gathered and organized into batches. \ding{203}~Entity mentions are extracted and resolved into an identity lattice that captures artifacts at varying specificity. \ding{204}~Dependency edges are constructed against the resolved lattice, with cross-source reconciliation and evidence grounding. The process repeats recursively upstream.}
    \label{fig:pipeline}
\end{figure}

With the dependency graph semantics fixed, \system{} recovers the graph from public release evidence. The system uses a staged agentic pipeline built around two core principles. First, discovery is separated from normalization: the system initially preserves local mentions and descriptions, then resolves them only after cross-source context is available. Second, every dependency claim must be grounded in source evidence and validated before it enters the graph. Our current implementation uses Claude Code~\cite{claude_code_docs} as the agentic harness.

\paragraph{Phase 1: Source gathering.}
Dependency evidence is scattered across many artifacts, and no single source type captures the full dependency structure of a modern LLM release. A technical report may describe the training pipeline without naming the exact dataset versions used in code; a dataset card may identify an upstream generator absent from the paper; and repository configuration files may reference mixtures never described in prose.

\system{} therefore begins from a target release and gathers its official public artifacts, including technical reports, model and dataset cards, code repositories, release blogs, and linked upstream artifacts. We restrict the system to official sources to avoid relying on unverified third-party claims. Because undocumented dependencies fall outside this evidence scope, the recovered graph should be interpreted as a lower bound on the true dependency structure.

To manage long contexts, the collected artifacts are organized into topically coherent batches, such as all resources associated with a model family, training stage, or dataset mixture.

\paragraph{Phase 2: Entity discovery and resolution.}
In the next phase, the system identifies dependent model and dataset artifacts from each source batch, recording mentions exactly as they appear in the source together with supporting evidence spans.

As discussed in \S\ref{subsec:task-def}, a key challenge is entity ambiguity: the same artifact is often referenced at different levels of specificity across sources (e.g., ``Olmo 3 32B,'' vs. \texttt{Olmo-3-32B-Think} vs.\ \texttt{Olmo-3-1125-32B}), while datasets frequently appear under subsets, derived variants, or internal names. The system therefore maps extracted mentions into an \emph{identity lattice} defined in \S\ref{subsec:task-def}, e.g., \texttt{\{family:Olmo 3, size:32B, stage:Think\}}.

For Hugging Face artifacts, deterministic metadata retrieval validates canonical URLs and recursively follows parent, subset, and derivation relationships when available. This enables ambiguous dataset names and internal slugs to be mapped to the correct public artifacts. Identity collisions, nonexistent artifacts, and inconsistent resolutions are flagged before dependency construction, preventing these errors from propagating into the dependency graph.

\paragraph{Phase 3: Dependency construction and reconciliation.}
With artifact identities resolved, the goal of the next phase is to transform evidence scattered across sources into a unified, evidence-grounded dependency graph. The system re-reads each source batch against the resolved identity lattice. A lattice search tool allows the agent to resolve mentions to the most specific node supported by the evidence and emit structured operations. Each edge records upstream and downstream artifacts, a free-form role description, a coarse dependency label, and supporting source anchors. Edges that reference unresolved artifacts or lack supporting evidence are rejected.

A key challenge is reconciling relationships described across multiple sources. The same dependency may be referenced at different levels of specificity or through complementary pieces of evidence. \system{} therefore merges claims that describe the same relationship, preserving the most specific artifact identities available while aggregating supporting evidence. When sources disagree, e.g., by assigning a dependency to different sibling artifacts or describing incompatible roles, the system flags the case for review rather than silently resolving it. Flagged cases are examined by a dedicated audit stage, with unresolved cases escalated to human annotation.

The resulting dependency operations trace paths from every discovered artifact back to the target release, yielding a unified dependency graph.

\paragraph{Recursive expansion.}
The preceding phases recover the local dependency neighborhood described by the target release's own public artifacts. However, many important dependencies are only visible by recursively tracing upstream artifacts. For example, a target release may identify an OCR model used in data preprocessing, but the target release typically will not document the OCR model's own base checkpoint, training data, or synthetic-data generators. 

To recover this transitive structure, \system{} recursively applies the same pipeline to discovered upstream artifacts. Each newly discovered model or dataset can become a tracing target whose official artifacts are gathered, resolved, related, and reconciled into the global graph. Users may choose breadth-first search for maximal coverage, depth-first search for targeted investigation of particular chains, or beam search to expand the top-$K$ structurally central ancestors at each depth. This recursive expansion is what turns scattered one-hop evidence into an auditable dependency graph over the broader model and dataset ecosystem.

\section{Evaluation}
\label{sec:eval}

Evaluating \system{} is inherently challenging because complete ground-truth dependency graphs do not exist. We initially attempted to construct small-scale human-annotated graphs for evaluation, but quickly found this approach impractical:
even for a single model, experts (authors of this work) spent many hours tracing dependencies yet still failed to produce a reasonably exhaustive graph.

We therefore evaluate systems by the number of \emph{verified} dependency relationships they recover. For each recovered relationship, we perform post-hoc verification  using the cited evidence produced by \system{}. As the number of dependencies to verify is exponentially large, we use Claude Sonnet 4.6~\cite{anthropicsonnet46} with web search to assist this process: the verifier reads the cited evidence URLs, independently corroborates them, and returns a JSON verdict of \emph{verified}, \emph{refuted}, or \emph{unclear}, with a short explanation. Only relationships judged \emph{verified} are counted in the evaluation metric; refuted and unclear relationships are excluded from the verified-dependency count.

We evaluate four LLM releases with extensive public artifacts---papers, model and dataset cards, codebase, etc.: Olmo 3, Nemotron 3 Super~\cite{nvidia2026nemotron3superopen}, DR Tulu~\cite{DBLP:journals/corr/abs-2511-19399}, and SmolLM3~\cite{bakouch2025smollm3}. These span fully open releases, industrial open-recipe models, post-training pipelines, and compact fully open models.

We compare against general-purpose LLM and agent baselines prompted to recover the same dependency graph from public evidence: OpenAI GPT-5.5 Pro~\cite{openai2026gpt55}, OpenAI GPT-5.4 Pro~\cite{openai2026gpt54}, Claude Code with a single-prompt configuration (denoted as CC-single), and ChatGPT Deep Research~\cite{openaidr}. CC-single receives the same high-level task specification, but not \system{}'s staged decomposition into recursive discovery, extraction, canonicalization, evidence grounding, and validation. The remaining baselines are given the same target model and recursive graph-construction instructions. Additional prompting details appear in \S{\ref{app:baseline-prompt}}. 

Because \system{} produces a single entity-resolved graph across investigations, we report three scopes: \emph{depth-1}, which counts only relationships whose subject is the target model itself; \emph{unbounded}, which counts every relationship uncovered during $T$'s recursive investigation that is also forward-reachable from $T$ in the merged graph; and \emph{BFS reachability}, which additionally counts relationships forward-reachable from $T$ through the merged graph, including findings produced by separate sub-investigations of upstream artifacts. We attribute a relationship to $T$ in the unbounded scope when its cited evidence was gathered during $T$'s investigation and its subject node is forward-reachable from $T$ in the merged graph; the unbounded scope is the best comparison against baselines, while BFS reachability is reported for completeness over the recovered graph and allows an edge to be attributed to multiple targets that forward-reach it. These scopes are strictly nested: depth-1 $\subset$ unbounded $\subset$ BFS reach. Full details on our evaluation protocol can be found in \S{\ref{app:eval-protocol}}.

\begin{table*}[t]
        \caption{Evaluation by target model. Each cell reports the number of verified dependency edges recovered for that
        target. CC-single denotes the single-prompt Claude Code baseline. \system{} (depth-1) considers only relations whose subject is the target's
        canonical identifier; \system{} (unbounded) additionally counts relations discovered during $T$'s investigation that are forward-
        reachable from $T$ in the merged graph; \system{} (BFS reach.) counts every relation forward-reachable from $T$. Bolded values are
        the column-best across all rows.}
        \label{tab:eval-by-model}
        \vspace{.3em}
        \centering
        \small
        \setlength{\tabcolsep}{7.0pt}
        \begin{tabular}{lrrrrr}
            \toprule
            & \textbf{Olmo 3}
            & \textbf{Nemotron 3}
            & \textbf{DR Tulu}
            & \textbf{SmolLM3}
            & \textbf{Total} \\
            \midrule
            \multicolumn{6}{l}{\emph{Baselines}} \\
            GPT-5.5 Pro
                & 59 & 156 & 46 & 53 & 314 \\
            GPT-5.4 Pro
                & 87 & 75 & 44 & 77 & 283 \\
            CC-single
                & 91 & 78 & 37 & 69 & 275 \\
            ChatGPT Deep Research
                & 37 & 70 & 28 & 36 & 171 \\
            \midrule
            \multicolumn{6}{l}{\emph{Ours}} \\
            \system{} (depth-1)
                & 182 & 237 & 42 & 23 & 484 \\
            \system{} (unbounded)
                & 305 & 613 & 44 & 98 & 1{,}060 \\
            \system{} (BFS reach.)$^{\dagger}$
                & \textbf{481} & \textbf{1{,}023} & \textbf{62} & \textbf{127} & \textbf{1{,}654} \\
            \bottomrule
        \end{tabular}

        \vspace{0.3em}
        {\footnotesize $^{\dagger}$Per-target BFS counts may overlap; the Total reports the union of verified forward-reachable edges
        across the four targets.}
    \end{table*}

    \paragraph{Results.}

    Table~\ref{tab:eval-by-model} shows that all single-prompt baselines recover a comparable number of verified dependencies (171--314).
    In
    contrast, \system{} recovers substantially more verified dependencies. Even in the most conservative depth-1 scope, \system{} recovers
    484 verified relationships, exceeding the strongest baseline by 54\%. Under the unbounded scope, \system{} recovers 1{,}060 verified
    relationships---more than 3$\times$ the strongest baseline. Under BFS reachability, the recovered graph contains 1{,}654 verified
    forward-reachable relationships across the four targets. The gains are largest for Olmo~3 and Nemotron~3 Super, whose multi-stage data
    pipelines involve many intermediate datasets, generators, filters, and post-training artifacts.

\section{Findings}\label{sec:findings}

In this section, we analyze the graph recovered by \system{} and highlight findings that would be difficult to surface from individual sources alone. One advantage of representing recursive dependencies as a structured graph is that audit questions become executable queries rather than manual document searches. We can issue graph/SQL-style queries over the recovered graph. The findings below were surfaced through this query-driven audit workflow: structured queries identify candidate risk patterns, and the attached source anchors make those candidates verifiable. 

  \subsection{Quantitative Findings}

  Across the analyzed releases, \system{} recovers 2{,}526 artifact nodes, 9{,}112 dependency edges, and 36{,}187 evidence anchors. 1{,}443
  nodes are datasets, 1{,}083 are models, and many model dependencies enter through generated, filtered, or rewritten data rather than
  through checkpoint inheritance. Per-target ancestor counts and maximum depths are reported in \S{\ref{appendix:graph-scale}}.

  \begin{table*}[t]
        \centering
        \myfontsize
        \setlength{\tabcolsep}{3pt}
        \renewcommand{\arraystretch}{1.2}
        \caption{Verified dependency edges grouped by audit role, broken down by target investigation and upstream artifact type. Each cell
        reports the verified edge count for that (role, target, upstream-type). Edges are
    attributed via forward BFS reachability. The same edge may be reached from multiple targets, so per-target columns sum to more than the
    Total column, which reports the union of verified reachable edges. Direct-role edges
    materially shape weights or training data; indirect-role edges influence development without entering training.}
        \label{tab:relation-roles-granular}
        \begin{tabular*}{\textwidth}{@{\extracolsep{\fill}}lrrrrrrrrr@{}}
          \toprule
          \textbf{Audit role}
          & \multicolumn{2}{c}{\textbf{Olmo 3}}
          & \multicolumn{2}{c}{\textbf{Nemotron 3}}
          & \multicolumn{2}{c}{\textbf{DR Tulu}}
          & \multicolumn{2}{c}{\textbf{SmolLM3}}
          & \textbf{Total} \\
          \cmidrule(lr){2-3} \cmidrule(lr){4-5} \cmidrule(lr){6-7} \cmidrule(lr){8-9}
          & Model & Data & Model & Data & Model & Data & Model & Data & \\
          \midrule
          \multicolumn{10}{@{}l}{\textbf{\emph{Direct dependencies}}} \\
          Training inputs                          & 0  & 172 & 0   & 547 & 0  & 11 & 0  & 87 & 813 \\
          Upstream operations on training data     & 85 & 11  & 253 & 13  & 7  & 1  & 7  & 6  & 350 \\
          Weight-level model lineage               & 4  & 0   & 20  & 0   & 3  & 0  & 3  & 0  & 28 \\
          \textbf{Direct total}                    & \textbf{89} & \textbf{183} & \textbf{273} & \textbf{560} & \textbf{10} & \textbf{12} &
          \textbf{10} & \textbf{93} & \textbf{1{,}191} \\
          \midrule
          \multicolumn{10}{@{}l}{\textbf{\emph{Indirect dependencies}}} \\
          Evaluation / ablation                    & 6 & 163 & 20 & 122 & 8 & 28 & 3 & 10 & 360 \\
          Methodology / audit influence            & 8 & 32  & 24 & 24  & 1 & 3  & 7 & 4  & 103 \\
          \textbf{Indirect total}                  & \textbf{14} & \textbf{195} & \textbf{44} & \textbf{146} & \textbf{9} & \textbf{31} &
          \textbf{10} & \textbf{14} & \textbf{463} \\
          \midrule
          \textbf{Grand total}                     & \textbf{103} & \textbf{378} & \textbf{317} & \textbf{706} & \textbf{19} & \textbf{43}
          & \textbf{20} & \textbf{107} & \textbf{1{,}654} \\
          \bottomrule
        \end{tabular*}

        \vspace{0.3em}
        \footnotesize Of the 9{,}112 edges in the merged graph, 7{,}458 are not forward-reachable from any of the four targets (siblings, predecessors, and parallel-family artifacts not graph-downstream of any seed), and are excluded from this table.
    \end{table*}

  \begin{table}[t]
        \centering
        \myfontsize
        \setlength{\tabcolsep}{6pt}
        \caption{Audit roles and their constituent relation types. Verified edge counts per role and target are reported in
        Table~\ref{tab:relation-roles-granular}.}
        \label{tab:relation-roles}
        \begin{tabular}{@{}ll@{}}
        \toprule
        Audit role & Constituent relations \\
        \midrule
        \multicolumn{2}{@{}l}{\textbf{\emph{Direct roles}}} \\
        Training inputs & \texttt{trained\_on} \\
        \addlinespace[2pt]
        Upstream operations on training data &
          \begin{tabular}[t]{@{}l@{}}
          \texttt{generated\_by}, \texttt{filtered\_by}, \texttt{transformed\_by}, \\
          \texttt{embedded\_by}, \texttt{decontaminated\_against}, \texttt{composed\_from}
          \end{tabular} \\
        \addlinespace[2pt]
        Weight-level model lineage &
          \begin{tabular}[t]{@{}l@{}}
          \texttt{trained\_from}, \texttt{merged\_from}, \\
          \texttt{quantized\_from}
          \end{tabular} \\
        \midrule
        \multicolumn{2}{@{}l}{\textbf{\emph{Indirect roles}}} \\
        Evaluation / ablation &
          \begin{tabular}[t]{@{}l@{}}
          \texttt{used\_for\_evaluation},
          \texttt{used\_for\_ablation}
          \end{tabular} \\
        Methodology / audit influence & \texttt{inspired\_by} \\
        \bottomrule
        \end{tabular}
    \end{table}

  Table~\ref{tab:relation-roles-granular} breaks down the verified edges per target via forward BFS reachability, grouped by the audit
  roles defined in Table~\ref{tab:relation-roles}: an edge is attributed to target $T$ if its subject node is forward-reachable from $T$ in
  the merged graph. Because the same edge may be reached from multiple targets, per-target columns sum to more than the Total column, which
  counts each verified edge once. Evidence is fragmented across source classes: most operations are supported by only one source class, so
  analyses restricted to papers, model cards, or code alone would miss many dependencies (see Table~\ref{tab:source-type-distribution} in
  the Appendix for source-type statistics). Verified edges are dominated by direct dependencies (1{,}191 edges, 72.0\%) --- artifacts that
  materially enter the target's weights or training data --- with indirect dependencies accounting for the remainder (463 edges, 28.0\%)
  --- artifacts that shape development through evaluation, ablation, methodology borrowing. Within the direct dependencies, upstream models
  more often shape downstream systems through data operations than through weight lineage: generation, filtering, transformation,
  embedding, and decontamination account for 350 verified edges (21.2\%), compared with 28 (1.7\%) for direct checkpoint lineage.

  \begin{table}[t]
    \centering
    \myfontsize
    \setlength{\tabcolsep}{5pt}
    \caption{Internal vs.\ external dependencies per target. An edge is \emph{internal} if its upstream artifact shares the target's
    organization (e.g., Ai2 artifacts for Olmo 3 and DR Tulu, NVIDIA for Nemotron 3, HuggingFace for SmolLM3), and \emph{external}
    otherwise. Counts are verified BFS-reachable edges per target; an edge reached from multiple targets is counted in each.}
    \label{tab:internal-external}
    \begin{tabular}{@{}lrrrrrrr@{}}
    \toprule
    & \multicolumn{2}{c}{\textbf{Internal}} & \multicolumn{2}{c}{\textbf{External}} & & & \\
    \cmidrule(lr){2-3} \cmidrule(lr){4-5}
    Target & Model & Data & Model & Data & Int.\ total & Ext.\ total & Ext.\ \% \\
    \midrule
    Olmo 3        & 13 & 106 & 90  & 272 & 119 & 362 & 75.3\% \\
    Nemotron 3    & 40 & 203 & 277 & 503 & 243 & 780 & 76.2\% \\
    DR Tulu       & 2  & 9   & 17  & 34  & 11  & 51  & 82.3\% \\
    SmolLM3       & 3  & 28  & 17  & 79  & 31  & 96  & 75.6\% \\
    \bottomrule
    \end{tabular}
    \end{table}

    \paragraph{Internal vs.\ external dependencies.}
    We additionally classify each verified BFS-reachable edge as \emph{internal} if its upstream artifact shares the target's
    organization (e.g., \texttt{allenai/*} artifacts for Olmo 3 and DR Tulu, \texttt{nvidia/*} for Nemotron 3 Super, \texttt{HuggingFaceTB/
    *} for SmolLM3) and \emph{external} otherwise. Table~\ref{tab:internal-external} reports this split. Across all four targets, external
    dependencies dominate: 75--82\% of verified edges come from outside the target's organization. Olmo 3 reaches 90 external models (e.g.,
    \texttt{openai/gpt-4.1}, \texttt{Qwen/Qwen3-32B}, \texttt{Qwen/QwQ-32B}) versus 13 internal Ai2 models (e.g., \texttt{allenai/OLMo-2},
    \texttt{allenai/wildguard}); the dataset-level ratio is similar (272 external vs.\
    106 internal). OpenAI and Qwen are the most depended-on external organizations across all four targets.

\subsection{Qualitative Findings}

  We next highlight qualitative findings surfaced by \system{}. These findings were difficult to recover from any single source: they
  require joining papers, model cards, dataset cards, code, and recursively linked upstream artifacts. To the authors' best knowledge,
  these are either only known to a small number of experts or were not known even to authors of the original work based on our follow-up
  conversations with them. The examples below fall into six recurring audit patterns:

  \begin{FindingsBox}{Key audit patterns}
    \footnotesize
    \begin{tabularx}{\linewidth}{@{}X X@{}}
    \findingtile{Multi-hop upstream models}
    {Dependencies can be hidden behind intermediate datasets, filters, classifiers, teachers, and tools.}
    &
    \findingtile{Training--evaluation coupling}
    {Benchmarks can reappear as prompts, splits, validation environments, or synthetic expansions.}
    \\[5pt]

    \findingtile{License-relevant paths}
    {Synthetic-data and annotation pipelines may carry license or terms-of-use implications from upstream models.}
    &
    \findingtile{Model-mediated selection}
    {Generators, judges, filters, and reward models decide which examples are created, kept, or rewarded.}
    \\[5pt]

    \findingtile{Code-level provenance}
    {Training scripts, YAMLs, and data-prep code reveal dependencies hidden by paper/card summaries.}
    &
    \findingtile{Mitigations and hygiene}
    {The graph also surfaces good practices: permissive-teacher swaps, filtering, and decontamination.}
    \end{tabularx}
    \end{FindingsBox}

  \paragraph{Multi-hop upstream models.} A core advantage of recursive tracing is that it finds upstream models that never appear in the
  final model's own paper or card. These models are often hidden behind intermediate datasets, filters, classifiers, teachers, or tools, so
  they only become visible after following several evidence-backed hops.

  \begin{itemize}[leftmargin=14pt, topsep=1pt, itemsep=2pt]

      \item \textbf{DR Tulu depends on Claude-generated ScholarQA trajectories.} DR Tulu's main paper states that
      their training data is generated by OpenAI models, but the Appendix additionally references Ai2 ScholarQA~\cite{singh-etal-2025-ai2}
      trajectory data used to create SFT data. Following the ScholarQA implementation reveals that its generation pipeline uses Claude
      Sonnet 3.7 by default, finding a hidden dependency chain {Claude Sonnet 3.7}~\cite{anthropicsonnet37} $\rightarrow$ {Ai2 ScholarQA} $
      \rightarrow$ {DR Tulu}.

      \item \textbf{Olmo 3 RL-Zero inherits Qwen2.5-Coder through its data-construction chain.} Olmo 3 RL-Zero models are trained on Dolci
      RL-Zero mixtures derived from earlier Olmo 3 checkpoints and code-oriented data construction. Tracing those intermediate artifacts
      back to Olmo 3's midtraining pipeline surfaces Qwen2.5-Coder-32B-Instruct~\cite{DBLP:journals/corr/abs-2409-12186} as an upstream
      model used to transform code data. This dependency is not stated in the downstream RL-Zero cards.

  \end{itemize}

  \paragraph{Training--evaluation coupling.} Another finding is the pervasive \emph{structural coupling} between training pipelines and
  evaluation benchmarks. These are often not direct cases of test-set leakage, but more subtle recursive relationships in which benchmark
  prompts, training splits, auxiliary resources, or validation environments are transformed into training artifacts while the same
  benchmark family remains an evaluation target. Such couplings are rarely visible from any single artifact and often evade conventional
  decontamination because the relevant connections emerge only after recursively tracing dependencies across multiple hops.

  \begin{itemize}[leftmargin=14pt, topsep=1pt, itemsep=2pt]

      \item \textbf{Olmo training mixes reuse benchmark-derived data.} \emph{IFEval}: Olmo 3 IF-RLVR prompts have their constraints sampled
      from IFEval~\cite{DBLP:journals/corr/abs-2311-07911} and IFBench-Train~\cite{DBLP:journals/corr/abs-2507-02833}, while IFEval also
      appears as an evaluation target in the same release. \emph{GSM8K}: Olmo 2's Dolmino-100 anneal mix lists the
      GSM8K~\cite{DBLP:journals/corr/abs-2110-14168} train splits directly in \texttt{dolmino100.txt} plus a TinyGSM~\cite{DBLP:journals/corr/abs-2312-09241}-style synthetic expansion (\texttt{gsm\_MIND/clean\_stop}); GSM8K is reported as an evaluation target across
      both Olmo 2~\cite{DBLP:journals/corr/abs-2501-00656} and Olmo 3. This creates a structural train/eval coupling that is invisible from
      the model paper alone.

      \item \textbf{Nemotron products disagree on how to handle SWE-Bench-Verified.} Nemotron-3-Super repurposes data that was derived from
      SWE-Bench-Verified~\cite{swebenchverified} for RL training via \texttt{nvidia/Nemotron-RL-Agentic-SWE-Pivot-v1}, then reports SWE-Bench-Verified as a headline evaluation. Nemotron-Cascade~\cite{DBLP:journals/corr/abs-2512-13607} takes the opposite approach: it
      removes SFT examples whose source repositories appear in SWE-Bench-Verified before evaluating on the benchmark. Thus, the same
      benchmark is treated as a training signal in one Nemotron release and as contamination risk in another.

      \item \textbf{Popular benchmarks repeatedly play both training-side and evaluation roles.} Aggregating roles per benchmark, \system{}
      reveals that several benchmarks appear both as evaluation targets and as training-side artifacts: GSM8K (25 evaluation edges, 43
      training edges), MMLU~\cite{DBLP:conf/iclr/HendrycksBBZMSS21} (39 evaluation, 14 training), GPQA~\cite{DBLP:journals/corr/abs-2311-12022} (45 evaluation, 9 training), MATH~\cite{DBLP:conf/nips/HendrycksBKABTS21} (31 evaluation, 30 training), IFEval (27 evaluation,
      18 training), and SWE-bench Verified (2 evaluation, 10 training).

  \end{itemize}

  \paragraph{License-relevant paths.} Potential license or terms-of-use implications are not always visible from the license attached to a final dataset or model. Prior work
  argues that dataset legal risk cannot be assessed from license terms alone and instead requires tracing redistribution and lifecycle
  history~\cite{kim2025trustlicensesseedataset}; supply-chain audits similarly find that license labels can function as weak metadata
  signals when the underlying compliance payload is missing or fails to propagate~\cite{jewitt2026permissivewashingopenaisupply}.

  \begin{itemize}[leftmargin=14pt, topsep=1pt, itemsep=2pt]
      \item \textbf{Major model families directly shape many downstream releases.} Even counting only one-hop training-side dependencies, a
      small set of model families dominate: Qwen touches 167 downstream artifacts through 552 edges, Llama touches 157 artifacts through
      264 edges, GPT-4~\cite{gpt4} touches 65 through 125 edges, and DeepSeek touches 81 through 162 edges.

      \item \textbf{SmolLM3's FineMath data traces back to Llama-generated annotations.\footnote{The Llama 3 Community License Agreement includes the clause: \emph{``You will not use the Llama Materials or any output or results of the Llama Materials to improve any other large language model (excluding Meta Llama 3 or derivative works thereof).''} Whether classifier-training annotations constitute ``output or results of the Llama Materials'' used to ``improve'' SmolLM3 is an open interpretive question, but the dependency chain (Llama-3-70B-Instruct $\rightarrow$ educational-value annotations $\rightarrow$ finemath-classifier $\rightarrow$ FineMath $\rightarrow$ SmolLM3) illustrates the kind of multi-hop path with potential license implications that manual auditing is unlikely to catch.}} The SmolLM3 model card lists FineMath~\cite{DBLP:journals/corr/abs-2502-02737} as a pretraining source, and the FineMath card
      describes classifier-based filtering of mathematical web data. However, the classifier card finds that \texttt{finemath-classifier}
      was trained on Llama-3-70B-Instruct educational-value annotations. Thus, SmolLM3's pretraining data is shaped by a Llama-generated
      annotation pipeline even though the SmolLM3 card does not name Llama as an upstream source.

  \end{itemize}

   \paragraph{Model-mediated selection.} Many upstream models do not appear as explicit training datasets. Instead, they generate, judge,
   filter, score, or reward candidate data --- their outputs and preferences shape which
    examples are created, kept, or rewarded, even when no weights are copied. Moreover, the graph shows that this influence is highly
    concentrated: the same upstream model families recur as generators, filters, and judges across
    many releases, an ecosystem-level dependency pattern not visible from individual papers, which often name only local datasets or
    summarize synthetic-data construction at a family level.

    \begin{itemize}[leftmargin=14pt, topsep=1pt, itemsep=2pt]
        \item \textbf{Qwen3-32B judges Olmo 3 RL data.} The Olmo 3 paper notes the use of Qwen3-32B~\cite{DBLP:journals/corr/abs-2505-09388} as an LM judge, while training scripts show that this judge is used for RL-Zero prompts without verifiable ground truth. Thus, Qwen3-32B's preferences can shape which RL examples are retained or rewarded.

        \item \textbf{Nemotron-3-Super uses a Qwen-bootstrapped reward model.}
        The generative reward model used in Nemotron-3-Super RLHF identifies Qwen3-235B-A22B-Thinking-2507 as the foundation model and
        names preference-data sources including \texttt{nvidia/HelpSteer3}~\cite{wang-etal-2025-helpsteer3} and commercially-friendly
        subsets of \texttt{lmarena-ai/arena-human-preference-140k}. The resulting chain---Qwen3 foundation $\rightarrow$ GenRM judge $
        \rightarrow$ Nemotron-3-Super---requires the Nemotron paper, the GenRM card, and the underlying preference-data card to assemble.

        \item \textbf{Nemotron-PrismMath routes Qwen-generated math data into Nemotron training.} Nemotron-3-Super and Nemotron-3-Nano
        variants train on \texttt{nvidia/Nemotron-PrismMath}~\cite{DBLP:journals/corr/abs-2505-20161}, whose card identifies Qwen2.5-72B-Instruct~\cite{DBLP:journals/corr/abs-2412-15115}, Qwen2.5-0.5B-Instruct, and DeepSeek-R1-Distill-Qwen-32B~\cite{DBLP:journals/nature/GuoYZSWZXZMBZY025} as generators for its problem--solution pairs.

        \item \textbf{A few model families dominate generator and judge roles.} Among generator, filter, transformer, and judge edges, the
        most-used upstream models include Qwen2.5-32B-Instruct (70 uses), DeepSeek-R1 (43),
        Llama-3.3-70B-Instruct (36), Qwen3-32B (33), and GPT-4.1 (32). This shows that synthetic-data construction concentrates around a
        small number of high-use upstream models, especially in post-training stages such as DPO, RLHF, and RLVR.
    \end{itemize}

  \paragraph{Code-level provenance.} Release cards and papers often summarize a model's data too coarsely to determine what was actually
  used in training. In several cases, the decisive evidence appears only in training scripts, YAML mixtures, or dataset-construction
  code.\footnote{Dependencies found only in code may be either genuinely ancillary (e.g., a preprocessing utility not considered
  part of the formal pipeline) or direct training dependencies that the paper simply omits; the graph surfaces both for audit, but this
  distinction matters when interpreting documentation gaps.}

  \begin{itemize}[leftmargin=14pt, topsep=1pt, itemsep=2pt]

      \item \textbf{Nemotron training blends are not fully specified until code fills in missing datasets.} Nemotron-Super-RL-Training-Blends includes placeholder rows for DAPO-Math~\cite{yu2026dapo} and Skywork-OR1~\cite{he2025skyworkopenreasoner1} contributions,
      requiring a separate \texttt{fill\_placeholders.py} script to fetch and restore upstream data before the blend is usable. Similar
      placeholder mechanisms appear in Nemotron-3-Nano~\cite{nvidia2025nemotron3nanoopen} and Nemotron-3-Nano-Omni~\cite{nvidia2026nemotron3nanoomni}, suggesting a recurring strategy in the Nemotron family.

      \item \textbf{SmolLM2's SmolTalk pipeline collapses legacy corpora and teacher/filter models into one SFT dataset name.} The
      SmolLM2~\cite{DBLP:journals/corr/abs-2502-02737} model card points to SmolTalk~\cite{DBLP:journals/corr/abs-2502-02737}-style SFT
      data, and the SmolTalk card describes broad components such as summarization, rewriting, and Magpie-style instruction data. However,
      the code reveals the concrete upstream construction path: the summarization branch loads CNN DailyMail~\cite{10.5555/2969239.2969428}
      and processes it through Qwen2.5-72B-Instruct, while the Magpie-Ultra branch generates instructions with Llama-3.1-405B-Instruct and
      filters them with Llama-3.1-8B-Instruct.

      \item \textbf{Olmo 3 DPO scripts reveal which models generated preference pairs.} The Olmo 3 paper describes DPO data at a high
      level, but the 32B-Instruct DPO training script names synthetic-pair datasets whose filenames encode their upstream generators and
      construction steps, including GPT-3.5~\cite{chatgpt}/GPT-4o~\cite{4o} preference-pair sources, multi-turn truncation, deduplication,
      and topic filtering.

  \end{itemize}

  \paragraph{Mitigations and hygiene.} Several examples show model developers taking care across multiple hops. These choices are scattered
  across documents, and the graph surfaces them as deliberate release practices rather than isolated notes.

  \begin{itemize}[leftmargin=14pt, topsep=1pt, itemsep=2pt]
      \item \textbf{Ai2 rebuilds upstream recipes with more permissive teachers.} Three Olmo 3 midtraining subsets share the same design
      pattern: CraneMath reimplements the SwallowMath~\cite{DBLP:journals/corr/abs-2505-02881} recipe using Qwen3 instead of Llama;
      CraneCode applies the same teacher swap to a SwallowCode~\cite{DBLP:journals/corr/abs-2505-02881}-style rewriting pipeline using
      Qwen2.5-Coder-32B-Instruct; and MegaMatt recreates a MegaMath~\cite{DBLP:journals/corr/abs-2504-02807}-style methodology with Qwen3.

      \item \textbf{Olmo 3 filters Llama-Nemotron data to avoid Llama-touched samples.} Olmo 3 includes Llama-Nemotron Post-Training data
      in its training mixture, but the retained reasoning samples are filtered to DeepSeek and Qwen samples that were not touched by Llama
      models.

      \item \textbf{Nemotron-Cascade SWE-SFT excludes repositories that appear in SWE-Bench Verified.}
      Where a typical model card simply claims ``decontamination was applied,'' the Nemotron-Cascade dataset card records the operational
      definition: SWE SFT instances whose source repository appears in \texttt{princeton-nlp/SWE-bench\_Verified} are dropped before SFT.
      This repository-level exclusion is stronger than typical text-level overlap removal.

  \end{itemize}

  These findings show that recursive model--model dependency graphs provide a strong auditing layer that can surface key issues previously
  opaque due to the complexity of modern LLM development. Further examples are in \S{\ref{appendix:findings}}. Because \system{} uses
  reported information from public artifacts, these findings should be interpreted as a lower bound on the true dependency structure.
  Undocumented uses of private models, unreleased data mixtures, or internal filtering pipelines may introduce additional dependencies that
  are not recoverable from public evidence.

\section{Conclusion}
We formalize recursive LLM dependency tracing as the task of constructing evidence-grounded dependency graphs over model and dataset artifacts, and present \system{}, an agentic system that recovers such graphs from public artifacts. We find that auditing modern LLM dependencies requires explicit graph semantics for what counts as a dependency, how heterogeneous pipeline roles should be represented, and how artifact identities should be reconciled across inconsistent sources. Across four public-artifact-rich LLM releases, \system{} recovers over a thousand verified dependency relationships, turning fragmented public evidence into structured graphs that support audit queries. These graphs surface issues difficult to identify from individual sources alone, including license-relevant multi-hop paths, structural train/evaluation coupling, mismatches between released artifacts and trained-on artifacts, and documentation inconsistencies. As LLM pipelines become increasingly recursive and model-mediated, transparency efforts must move beyond flat documentation toward disclosure schemas that explicitly represent dependency structure, evidence, and role semantics.

\section*{Limitations}
Our results should be interpreted in light of several limitations. First, \system{} reconstructs dependencies from publicly available artifacts and therefore cannot recover dependencies that are undocumented, proprietary, or otherwise inaccessible. As a result, the graphs we produce represent an evidence-grounded lower bound on the true dependency structure, with the gap likely largest for closed or partially disclosed releases.

Second, complete ground-truth dependency graphs do not exist, making absolute recall difficult to measure. Our evaluation therefore focuses on four well-documented LLM releases for which extensive supporting artifacts are available. While these models provide a challenging and realistic testbed, future work is needed to assess coverage across less thoroughly documented ecosystems.

Third, both \system{} and the automated verification pipeline rely on Claude Code. Although verification is constrained by explicit evidence grounding and deterministic validation rules, shared modeling biases could still affect both stages. Independent verification pipelines based on different model families would provide a stronger assessment of extraction quality.

Finally, our comparisons focus on fully automated baselines. We do not evaluate settings in which expert users iteratively guide a general-purpose agent through the dependency-tracing process. Such human-in-the-loop workflows could reduce the performance gap and would help disentangle the contribution of \system{}'s task decomposition from that of the underlying agent capabilities.

\section*{Acknowledgements}
We thank the creators of the model artifacts analyzed in this work (Olmo 3, Nemotron 3, DR Tulu, SmolLM 3, as well as intermediate models and datasets) for publicly releasing all training information that enabled this study.

We thank Kyle Lo, Noah Smith, Hanna Hajishirzi, Rishi Bommasani, the SM group members and Ai2 members for valuable discussion and feedback.

This work was supported in part by gifts from Ai2 and Apple.

\bibliographystyle{plain}
\bibliography{references.bib}

\clearpage
\appendix

\section{Evaluation Protocol Details}
  \label{app:eval-protocol}

  Our evaluation has three steps: we pool candidate relationships, verify each relationship against public evidence, and then attribute verified/refuted relationships to target models for reporting.

  \paragraph{Candidate relationship pooling.} For each target model, we collect the dependency relationships emitted by \system{} and all baselines. Each candidate relationship contains a subject artifact, an object artifact, a
  free-form dependency description, relation metadata, and supporting evidence URLs or excerpts.

  \paragraph{Verification.} Each candidate relationship is checked by Claude Sonnet 4.6 with web search. The BFS reachability scope and the full-graph aggregate use a separate audit by Claude Opus 4.7 with extended
  thinking and web search. The verifier reads the submitted evidence, checks cited excerpts or locations when available, and may search for
  independent public corroboration. 

  \paragraph{Attribution scopes for \system{}.} Unlike the baselines, \system{} produces one merged, entity-resolved graph across the four target investigations. We therefore need explicit rules for assigning each relationship
  in the merged graph back to one or more target models. We report three scopes.

  \emph{Depth-1} is the strict scope. A relationship is attributed to target $T$ only when its subject canonicalizes to $T$'s canonical identifier. Canonicalization lowercases the string and collapses non-alphanumeric runs to
  hyphens, while preserving any HuggingFace organization prefix. This scope is closest to the per-target outputs produced by single-prompt baselines: it only counts relationships where the target itself is the subject.

  \emph{Unbounded} combines worker provenance with forward graph reachability. A relationship is attributed to target $T$ if \emph{both} of the following hold:
  \begin{enumerate}[leftmargin=16pt, topsep=1pt, itemsep=0pt]
      \item \textbf{Provenance}: at least one of (i)~its subject canonicalizes to $T$, (ii)~at least one supporting anchor comes from $T$'s target-specific seed directory, or (iii)~at least one supporting anchor was produced by
  a worker whose outputs co-occur only with $T$'s seed directory.
      \item \textbf{Forward reachability}: the subject node is forward-reachable from $T$ in the merged graph by traversing edges in the direction subject $\to$ object.
  \end{enumerate}
  Seed directories contain the papers, model cards, dataset cards, READMEs, and training configs gathered specifically for one target investigation. Worker IDs identify short-lived extraction or relation-generation jobs; if a
  worker only co-occurs with one target's seed materials, we treat its outputs as part of that target's recursive investigation.

The provenance criterion captures edges actually discovered during $T$'s investigation, while the forward-reachability criterion ensures those discoveries are about artifacts in $T$'s dependency chain rather than context the worker happened to read about (e.g., sibling or predecessor artifacts in the same family). Together, the two criteria capture what each target investigation contributed to its own chain. A relationship may be attributed to multiple targets when its provenance spans multiple target investigations and it is forward-reachable from each. For completeness, we additionally report a broader graph-reachability scope, defined below, which counts every relation reachable from $T$ regardless of which investigation surfaced it. Under this nesting, depth-1 $\subset$ unbounded $\subset$ BFS reach.

\emph{BFS reachability} is the graph-reachability scope. A relationship is attributed to target $T$ if its subject node is forward-reachable from $T$ in the merged graph by traversing edges in the direction subject $\to$ object. An edge may be attributed to multiple targets when several seeds forward-reach it; the Total under this scope reports the union of reachable edges across the four seeds rather than their sum. This scope captures the full transitive dependency footprint of each target, including relationships uncovered by separate sub-investigations of upstream artifacts that share lineage with $T$. Sibling and predecessor artifacts that are not graph-downstream of any seed (e.g., earlier-generation models in the same family) fall outside this scope.

Under this unbounded attribution rule, $1{,}060$ of the $9{,}112$ relations in the merged graph are attributed to at least one of the four targets. Under the BFS reachability scope, $1{,}654$ unique edges in the merged graph are forward-reachable from at least one of the four targets.
  
\section{Baseline Prompt}
\label{app:baseline-prompt}

All baseline systems were given the same dependency-reconstruction prompt template, summarized in Table~\ref{tab:baseline-prompt-summary}. For each target model, we instantiated a \texttt{SUBJECT} block with the target's canonical identifier, display name, provider, release date, authoritative paper/repository/card URLs, scope note, and recursion depth. The prompt instructed baselines to recover both direct training-pipeline dependencies and indirect development dependencies, then emit a single JSON graph containing \texttt{subject}, \texttt{nodes}, and \texttt{edges}. It also specified the dependency scope, canonicalization rules, evidence requirements, recursion policy, and output validation checks. The full prompt template is provided in the supplementary artifacts as \texttt{baseline\_prompt.md}.

\begin{table}[h]
\centering
\small
\caption{Structure of the baseline prompt used for all comparison systems.}
\label{tab:baseline-prompt-summary}
\begin{tabular}{lp{0.66\linewidth}}
\toprule
Prompt component & Description \\
\midrule
Subject block &
Specifies the target model, authoritative sources, scope note, and maximum recursion depth. \\

Dependency scope &
Defines direct dependencies as artifacts that enter the training pipeline and indirect dependencies as artifacts that shape evaluation, ablation, or methodology decisions. \\

Node and edge schema &
Requires model/dataset nodes and evidence-grounded edges with \texttt{relation\_type}, \texttt{dependency\_kind}, free-form descriptions, and supporting evidence. \\

Canonicalization rules &
Standardizes model and dataset identifiers, aliases, provider names, and version conventions to improve alignment across systems. \\

Evidence requirements &
Requires each edge to include source URLs and explanations, with locations and excerpts recommended for verifiability. \\

Research protocol &
Instructs the baseline to read authoritative papers, model cards, dataset cards, repositories, release blogs, and upstream artifact documentation. \\

Recursion policy &
Requires recursive expansion of upstream artifacts until the depth cap is reached or authoritative documentation runs out. \\

Validation checklist &
Asks the baseline to check schema validity, node coverage, canonical-ID consistency, evidence grounding, recursion coverage, and exclusion of out-of-scope artifacts before emitting JSON. \\
\bottomrule
\end{tabular}
\end{table}

\section{Additional Quantitative Results}
  \label{appendix:quantitative-results}

  \subsection{Recovered Graph Scale by Target}
  \label{appendix:graph-scale}

  \begin{table}[h]
  \centering
  \small
  \caption{Scale and depth of recovered dependency graphs. Ancestors count unique transitive upstream artifacts reachable from each target
  model.}
  \label{tab:graph-scale-appendix}
  \begin{tabular}{lcc}
  \toprule
  Target model & Ancestors & Max depth \\
  \midrule
  Olmo 3 Instruct & 409 & 8 \\
  Olmo 3 Think & 329 & 8 \\
  Olmo 3 Base & 181 & 4 \\
  Nemotron-3-Super & 465 & 5 \\
  Nemotron-3-Nano-Base & 319 & 5 \\
  DR Tulu & 17 & 2 \\
  SmolLM3-Base & 78 & 3 \\
  \midrule
  \textbf{Aggregate graph} & \multicolumn{2}{c}{2,526 nodes, 9,112 edges, 36,187 anchors} \\
  \bottomrule
  \end{tabular}
  \end{table}

  Table~\ref{tab:graph-scale-appendix} shows the per-target ancestor counts and maximum depths. The variance in max depth across targets
  reflects pipeline complexity rather than recovery quality: SmolLM3-Base is a pretraining-only release whose recoverable lineage is
  dominated by web-scrape and filter steps (depth 3), whereas the Olmo 3 Instruct/Think families pass through midtraining, multiple SFT/
  DPO/RL stages, and judge-mediated synthetic-data construction (depth 8). Lineage size is also lower-bounded by what upstream
  documentation makes recoverable; releases with shallower public artifact trees (e.g., DR Tulu) yield correspondingly smaller ancestor
  sets.

  \begin{table}[t]
  \centering
  \small
  \caption{Source-type distribution of recovered edges. Single-source edges are supported by anchors from only one source class; multi-
  source edges are supported by anchors from multiple source classes.}
  \label{tab:source-type-distribution}
  \begin{tabular}{lcc}
  \toprule
  Source support & Count & Share \\
  \midrule
  Only Hugging Face cards & 669 & 7.3\% \\
  Only training code / scripts & 2,142 & 23.5\% \\
  Only PDFs / technical reports & 2,777 & 30.5\% \\
  Only release blogs / docs & 18 & 0.2\% \\
  Only other docs & 486 & 5.3\% \\
  Multiple source types & 3,020 & 33.1\% \\
  \midrule
  Total edges & 9,112 & 100\% \\
  \bottomrule
  \end{tabular}
  \end{table}

\section{Additional Qualitative Findings}
\label{appendix:findings}

The main text presents a curated subset of qualitative findings. This appendix gives additional examples that illustrate the same recurring audit patterns: hidden model-mediated data construction, benchmark exposure, release-artifact mismatches, and reproducibility issues.

\subsection{Hidden Model-Mediated Data Construction}

\begin{itemize}[leftmargin=14pt, topsep=1pt, itemsep=0pt]

\item \textbf{Nemotron-CC-v2.1 exposes Qwen3-rephrased Common Crawl as a downstream training source.} Nemotron-CC-v2.1 extends Common Crawl-derived data with synthetic rephrasing and translation using Qwen3-30B-A3B. User-facing model cards often describe this as synthetic Common Crawl, while dataset documentation exposes the upstream rephrasing model and the specific subset construction.

\item \textbf{Nemotron-3-Super uses training content generated in part by Nemotron-Nano-9B-v2.} The recovered graph identifies \texttt{nvidia/NVIDIA-Nemotron-Nano-9B-v2} as a generator for content used in Nemotron-3-Super training. This is a non-obvious family-internal data path: a smaller sibling model contributes training content for a larger downstream release.

\item \textbf{SmolLM2 is filtered by a Llama-3-derived reward model.} Following the SmolTalk pipeline reveals that \texttt{smol-magpie-ultra} examples are scored or filtered with \texttt{RLHFlow/ArmoRM-Llama3-8B-v0.1}. This makes a Llama-3-derived reward model an upstream filtering dependency for SmolLM2-Instruct, even though the dependency is not visible from the final model card alone.

\end{itemize}

\subsection{Benchmark Exposure and Decontamination}

\begin{itemize}[leftmargin=14pt, topsep=1pt, itemsep=0pt]

\item \textbf{LiveCodeBench appears both as RL validation and as an evaluation benchmark.}
A Nemotron training configuration references \texttt{livecodebench\_v5\_validation} as a validation split for a code-generation environment, while LiveCodeBench-style results are also reported as evaluation metrics. This raises a development-coupling question: validation benchmarks can influence checkpoint selection or training decisions even when they are not directly used as supervised training data.

\item \textbf{Olmo 3 RL-Zero applies a ``spurious-reward decontamination check'' against an enumerated benchmark suite.} Olmo 3-7B-RL-Zero-Math, -Code, and -Mix carry edges flagging spurious-reward decontamination checks against GSM8K, GPQA, AIME 2024, AIME 2025, Minerva Math, ZebraLogic, and Omega-500. These edges are scattered across model and dataset cards plus paper figure captions.

\item \textbf{Nemotron uses Qwen3-Embedding-0.6B as a decontamination filter.} The graph records that \texttt{Qwen/Qwen3-Embedding-0.6B} is used to encode candidate samples and remove examples with high cosine similarity to benchmark problems from HumanEval, MBPP, CRUXEval, and LiveCodeBench. Here the decontamination mechanism itself introduces a model dependency: the embedding model determines which data is retained or dropped.

\item \textbf{LLM360 MegaMath ships a per-benchmark decontamination receipt.} \texttt{LLM360/MegaMath-Llama-3.2-1B} and \texttt{LLM360/MegaMath-Llama-3.2-3B} each record 11 distinct \texttt{decontaminated\_against} edges: ASDiv, MATH, MathQA, MAWPS, MMLU-STEM, OCWCourses, SAT-Math, SVAMP, AIME 2024, AIME 2025, and AMC.
These edges are anchored to released decontamination artifacts such as \texttt{utils/decont\_utils/data/\{benchmark\}.jsonl}; where specified, the exclusion rule operates over question/answer fields using n-gram overlap.

\end{itemize}

\subsection{License and Release Hygiene}

\begin{itemize}[leftmargin=14pt, topsep=1pt, itemsep=0pt]

\item \textbf{Tulu 3 DPO data is generated by a wide cross-organization teacher panel.}
The Tulu 3 DPO mixtures include outputs from many upstream model families, including OpenAI, Anthropic, Mistral/Mixtral, 01-ai/Yi, MosaicML/MPT, and InternLM-family models. This makes DPO data a point where many license and terms-of-service regimes may accumulate, even when no single model card enumerates the full teacher panel.

\item \textbf{Olmo 3 releases both complete and redacted Dolma 3 variants, and its reproduction data changed after training.} Dolma 3 includes complete and redacted variants of the academic-document portion of the mix, including an olmOCR science-PDF slice. The reproduction dataset for Olmo-3-7B-1025 was later modified by replacing some redacted PDFs with \texttt{[REMOVED]}, which affects exact reproducibility. The graph surfaces this as a release-lineage issue rather than an isolated dataset-card note.

\end{itemize}

\subsection{Code and Configuration Reveal Details Hidden by Cards}

\begin{itemize}[leftmargin=14pt, topsep=1pt, itemsep=0pt]

\item \textbf{Per-stage placeholder percentages can be quantified by joining YAML and data-prep scripts.} Per-edge descriptions in the recovered graph quantify placeholder contributions: in \texttt{nvidia/Nemotron-RL-Super-Training-Blends}, DAPO-Math-17k contributes 1.36\% of the rlvr1 mix; Skywork-OR1-RL-Data contributes 5.44\% of the same mix; and the same DAPO-Math-17k contributes 0.10\% to the Nano blend.
These percentages are recoverable only by joining YAML mix-weight definitions with the upstream BytedTsinghua-SIA / Skywork dataset cards plus the \texttt{fill\_placeholders.py} data-prep script.

\item \textbf{SmolLM3 YAMLs reveal exact pretraining mixture details behind the card summary.} The SmolLM3 card summarizes pretraining with broad categories such as web, code, math, and multilingual data. The stage YAMLs instead pin individual per-language shards, including FineWeb2-HQ slices and Stack-Edu language buckets such as Python, Java, and Rust. Recovered SmolLM3 training edges also include exact mix weights.

\end{itemize}

\subsection{Methodology and Ecosystem-Level Dependencies}

\begin{itemize}[leftmargin=14pt, topsep=1pt, itemsep=0pt]

\item \textbf{SmolLM3 inherits methodology, not just data, from upstream models.}
Two recovered \texttt{inspired\_by} edges capture method transfer: SmolLM3 uses intra-document attention masking similar to Llama 3, and FineMath follows Qwen2.5-Math~\cite{DBLP:journals/corr/abs-2409-12122}'s 13-gram decontamination procedure. These examples illustrate a dependency class that is neither weight inheritance nor training-data reuse, but still shapes downstream model development.

\end{itemize}

\end{document}